\pdfoutput=1

\documentclass[11pt]{article}

\usepackage[]{EACL2023}

\usepackage{times}
\usepackage{latexsym}
\usepackage{siunitx}

\usepackage[T1]{fontenc}

\usepackage[utf8]{inputenc}

\usepackage{microtype}

\usepackage{inconsolata}

\usepackage{amsmath}
\usepackage{amssymb}
\usepackage{enumitem}
\usepackage{makecell}
\usepackage{dialogue}
\usepackage{amsbsy}
\usepackage{algpseudocode}
\usepackage{algorithm}
\usepackage{graphicx}

\title{Conversational Tree Search: A New Hybrid Dialog Task}
\author{Dirk V\"ath 
\And Lindsey Vanderlyn \\
University of Stuttgart, Germany \\
\texttt{\{vaethdk|vanderly|thang.vu\}@ims.uni-stuttgart.de}
\And Ngoc Thang Vu}

\begin{document}

\maketitle

\begin{abstract}
Conversational interfaces provide a flexible and easy way for users to seek information that may otherwise be difficult or inconvenient to obtain. 
However, existing interfaces generally fall into one of two categories: FAQs, where users must have a concrete question in order to retrieve a general answer, or dialogs, where users must follow a predefined path but may receive a personalized answer. 
In this paper, we introduce Conversational Tree Search (CTS) as a new task that bridges the gap between FAQ-style information retrieval and task-oriented dialog, allowing domain-experts to define dialog trees which can then be converted to an efficient dialog policy that learns only to ask the questions necessary to navigate a user to their goal.
We collect a dataset for the travel reimbursement domain and demonstrate a baseline as well as a novel deep Reinforcement Learning architecture for this task. 
Our results show that the new architecture combines the positive aspects of both the FAQ and dialog system used in the baseline and achieves higher goal completion while skipping unnecessary questions.
\end{abstract}

\section{Introduction}
Complex processes, e.g., healthcare, insurance, or travel reimbursement, can be challenging for users to navigate.
FAQ and task-oriented dialog systems can provide immediate support, which is especially helpful as questions in these areas can be time sensitive, e.g., whom to contact in case of a lost passport.
To support all types of users, however, such systems must be fast to satisfy those with more experience and thorough, so that those with less experience understand all steps needed to accomplish their goals.
Above all, however, accuracy is critical in such susceptible domains.

\begin{figure}[h]
    \centering
    \includegraphics[width=0.45\textwidth]{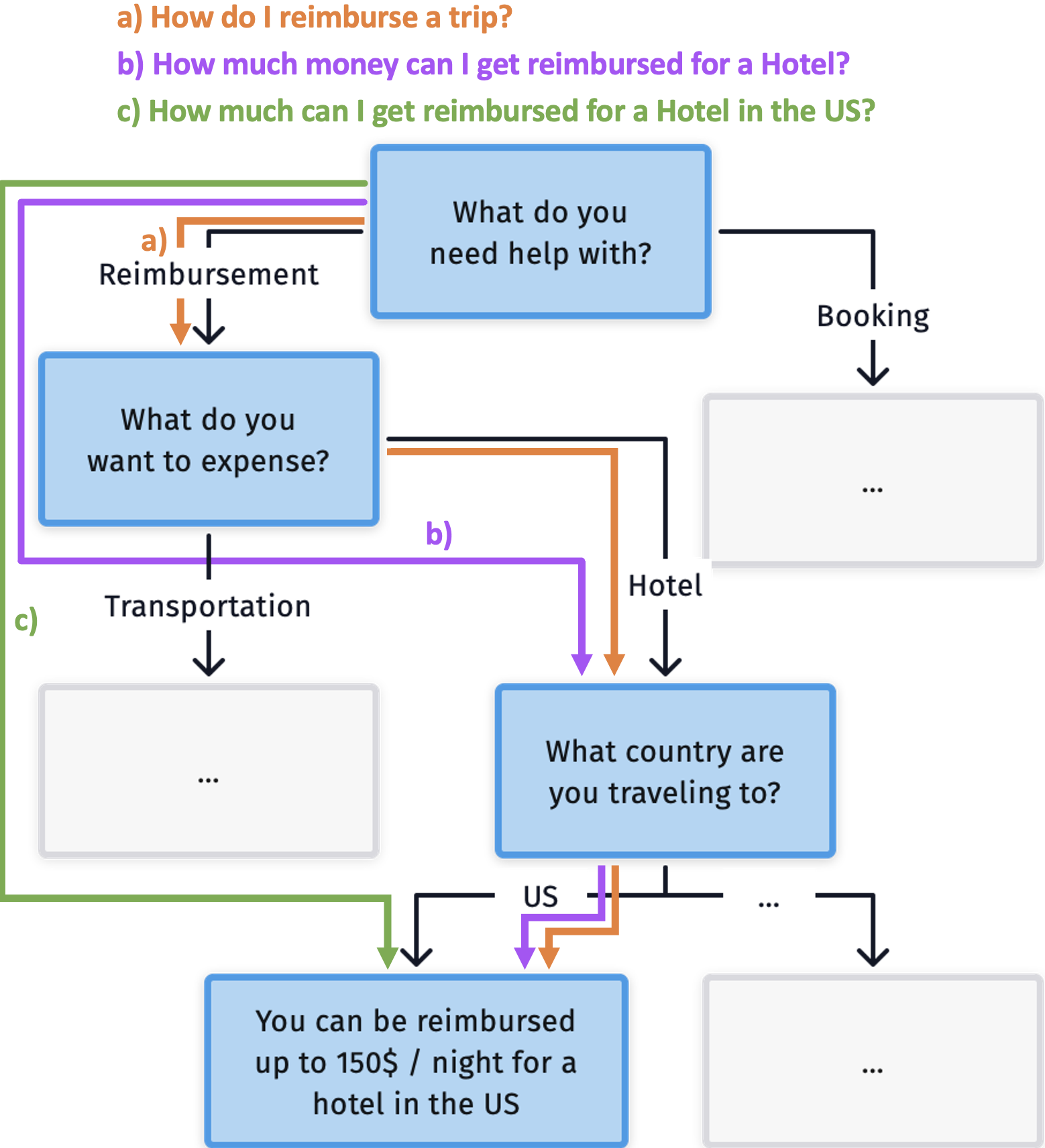}
    \caption{An example of the proposed task: Slice of a dialog tree (blue/gray nodes, black edges) showing how progressively more concrete questions could be answered. Question a) guiding a user with a general goal through the tree, b) asking only at nodes that need more clarification, and c) requiring no clarification and thus receiving a direct answer.}
    \label{fig:intro}
\end{figure}

FAQ systems can directly answer to user queries by matching them to predefined common question/answer pairs. 
It enables quick access to information while allowing subject experts to define system output and ensure it is factually correct \cite{10.1145/1066078.1066079}.
However, this approach does not allow for personalizing answers to a user's specific case, e.g., a specific user's per diem might depend on their destination and length of travel.
Including information for all cases in one FAQ answer would make it long and hard to understand, while adding FAQs for each case, would make retrieval challenging.
Additionally, as retrieval accuracy could be better \cite{thakur2021beir}, the top answer might be incorrect, but displaying multiple answers makes the search process longer and transfers the responsibility of selecting the correct answer to the user.
Finally, FAQ systems expect concrete information needs from users; but, new users may not be familiar enough with the complex process they wish to navigate in order to pose specific questions.

Dialog systems provide the opportunity for turn-based interaction, which can solve many of the shortcomings of FAQ systems (allowing process guidance, and shorter, more personalized answers).
While automatically learning dialog behavior can be quicker and allow for more flexibility than designing a handcrafted system, such dialog systems rely on large amounts of data \cite{raghu-etal-2021-end}, and their behavior can be challenging to control \cite{10.1145/3209978.3210183}. 
As such, handcrafted dialog systems have generally still been preferred in low-resource settings, where there is not enough data to train a machine learning approach \cite{zhang2020recent}, and in domains where it is essential that dialog designers can carefully control the system's behavior \cite{cohen2020back}. 
In these cases, it is common for non-technical experts to define the system behavior using a graphical interface \cite{shukla2020conversation}, resulting in a dialog tree.
However, the structure of such systems can be rigid to traverse, making it less suitable for experienced users who want a specific question answered as fast as possible, rather than having to navigate through the whole tree until they reach the answer they were looking for.

In trying to implement a dialog system for travel reimbursement, we confirmed the shortcomings of both FAQ- and dialog-based approaches through pilot studies. 
As we are unaware of any system at the time of writing that can handle both forms of information-seeking behavior, we propose a new task called Conversational Tree Search (CTS).
In this task, subject-experts can design arbitrarily complex dialog trees from which a dialog policy can be automatically trained to support both a guided dialog mode -- where users with a vague information need can be guided by questions each turn until they reach their answer -- and a free dialog mode -- where users with a concrete information need can receive an answer as efficiently as possible (see Figure \ref{fig:intro}). 
In this way, CTS offers a hybrid solution where experts can still carefully control dialog structure and system responses while also profiting from the flexibility of a machine learning policy that can navigate the tree efficiently, only asking questions necessary to clarify information needs or personalize answers.
Providing both dialog modes allows both novices and more experienced users to receive the correct level of support.

To solve our newly proposed task, we focus on the following research questions: \newline
\noindent \textbf{(RQ1)} How can we develop a dialog policy that can navigate an expert-designed dialog tree, supporting both users with a specific information need and those with only a vague information need?
\begin{itemize}[noitemsep,topsep=0pt]
    \itemsep0em 
    \item \textbf{(RQ1.1)} Can the policy learn to differentiate between both dialog modes?
    \item \textbf{(RQ1.2)} Can this policy learn only to ask questions that are necessary to reach the user goal?
\end{itemize}

\noindent \textbf{(RQ2)} How robust is such a system to noise added to the encoding of the user input? 
\newline
\noindent \textbf{(RQ3)} How well can such a system generalize to unseen utterances? 

In the process of answering these research questions, we make the following concrete contributions: 
1) We define a new task, CTS, and its evaluation criteria;
2) we collect and publish REIMBURSE, a low resource, real-world German language dataset for CTS in the domain of travel reimbursement, consisting of a dialog tree and user utterances;
3) we implement a domain-agnostic user simulator for CTS, using it to generate new dialogs based on the REIMBURSE data;
and 4) we design a novel Reinforcement Learning (RL) architecture and train an agent to achieve significant performance increases over the baseline system on noisy and unseen data.
Our data, simulator, and code are released\footnote{\label{footnote:code}\url{https://github.com/DigitalPhonetics/conversational-tree-search}} under an open source license to encourage more research into this task.

Our results show that an RL agent, based on our novel architecture, can learn to differentiate between both modes of interaction (guided and free) and skip unnecessary questions in free mode. 
Furthermore, the RL agent can significantly outperform the baseline system even in settings where Gaussian noise is applied to the user utterance encodings.
Finally, the RL agent outperforms the baseline on dialogs that are generated based on unseen user utterances.
Its success in this setting indicates an understanding of the user's text input, as the agent cannot exploit structural features, e.g., positional information.

\section{Definition of Conversational Tree Search Task}

The goal of this task is, given a dialog tree, to develop an interactive dialog system able to efficiently navigate a user to their goal by traversing the tree, and only outputting questions which are needed to clarify the user's intention and personalize the output accordingly.
In order to accomplish this, a policy has two types of actions it can choose at any node, either to $ASK$, which represents outputting the system text associated with that node, or to $SKIP$, which represents transitioning to a connected node without outputting the text at the current node. 

\subsection{Formal Task Description}
\begin{figure}[tb]
    \centering
    \includegraphics[width=0.3\textwidth]{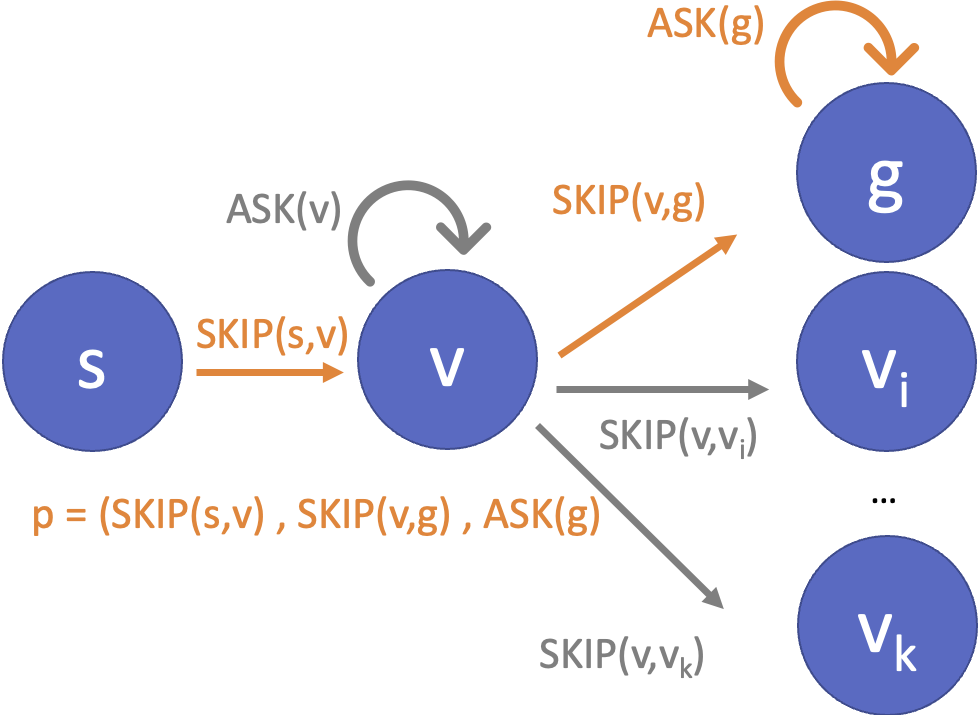}
    \caption{Example graph with dialog actions and path $p$ from start node $s$ to goal node $g$. As $ASK-g \in p$, this represents a dialog where the user reached their goal.}
    \label{fig:graph}
\end{figure}

Given is a dialog tree represented as (cyclic) directed Graph $G = (V,E)$,
where the outgoing edges of node $v$ include the self-transition $ASK(v)=(v, v)$, combined with the edges to  neighbours of $v$ (${v_i, \dots, v_k} \subseteq V$):
$E(v) = \{
    ASK(v), 
    SKIP(v,v_i), 
    \dots SKIP(v,v_k) 
\}$.
Here, $SKIP(v,v_i) = (v, v_i) $ is an edge to a direct neighbour $v_i$ of $v$.

Starting from a node $s \in V$,
the goal is to find a path $p = (ASK(s), SKIP(s,v_i), \dots, ASK(g))$ to a given user goal node $g \in V$ (see Figure \ref{fig:graph}) only described by user utterances $U$,
subject to $ ASK(g) \in p$ (the goal node should be presented to the user) while at the same time minimizing ${ |\{ ASK(v) | v \in V \wedge ASK(v) \in p \}| }$ (the amount of asked questions, i.e., perceived length to the goal, should be as few as possible).

Here, the edges $E$ also define the system's dialog actions, where $ASK(v)$ will output the text associated with the node $v$.
Depending on the node type, this can either output information to the user or extend the user utterance history $U$ as $U' = U \cup \{ u_{v}\} $ by asking a question.
The action $SKIP(v,v')$ will skip from $v$ to $v'$ without outputting anything to the user. 
Each node $v$ is associated with text representing a system utterance and a node type (e.g. question).
$SKIP$-actions are labeled by domain experts with a user answer prototype, which are used to compare user utterances to during dialog.
For more information on node and edge data, see Section \ref{sec:data}.

In order to address both user interactions (general exploration and specific questions), we define two goal settings within this task framework:

\paragraph{Guided Dialog}
Here, we define guided dialog as interactions where the user has a vague information goal and rather than posing a concrete question, would like to explore the information available in the dialog graph. 
Therefore, rather than having one static goal, the user decides on a new goal every turn, i.e., each turn their new goal will be to reach the node connected to the edge associated with the answer they give to the system question.
Thus, a guided dialog can be seen as having  turn-by-turn goals $(g, g', g'', \dots) \subset V$, meaning that from any node $g$, the next goal $g'$ is an immediate neighbour of $g$, i.e., $SKIP(g,g') \in p$.

\paragraph{Free Dialog}
Free dialog in contrast, considers the case that the user has expressed a concrete information goal, which may or may not require later clarification. 
In this form of dialog, rather than focusing on choosing the next node from the set of neighbours, the dialog system's goal is to help the user fulfill their information need as quickly as possible. To this end, if the system is not sure about an upcoming decision, it may choose to ask for relevant information, thereby increasing its understanding of the user's goal.
Thus, a free dialog has only one goal $g \in V$ and each turn serves to clarify the goal or skip closer to the answer.

\subsection{Evaluation Objectives}
We evaluate the path $p$ taken by the policies according to the following criteria:
\begin{enumerate}
    \item Task success, i.e., the goal node ($g \in V$) text is outputted to the user: $ASK(g) \in p$
    \item Skip ratio, i.e., the number of times two consecutive skip actions occur along the path ($(SKIP(v,v_i), SKIP(v_i,v_j)) \in p$) divided by the length of the path 

\end{enumerate}
For guided dialog, the objective is to maximize task success while minimizing the skip ratio.
In free mode, the objective is to maximize both task success and the skip ratio.

\section{Methods}

We develop and release\footnotemark[1] a baseline system, an RL-agent, and a domain-agnostic user simulator for training and evaluating dialog policies on the new CTS task.

\subsection{User Simulator}
The user simulator generates new dialogs for both interaction modes (with 50\% probability of starting in either mode), which can be used to train and evaluate an RL-agent.
The simulator is rules-based and each turn can respond to the dialog agent, either by asking a question (initial turn) or responding to a follow-up question (subsequent turns) using text from the available paraphrases (Appendix \ref{appendix:simulator}).
Dialogs are generated by choosing random goals and constraints.
As its behavior only depends on an interchangeable dialog tree, the simulator remains domain-agnostic.

To represent exploratory users (guided mode), the simulator randomly samples a neighboring node as a new goal after each node transition.
This mimics a user whose goal is only that the system correctly understands their current input and takes them to the associated next node.
To support user exploration of the domain, skipping over nodes should be avoided, so asking after skipping to a new node -- rather than skipping twice in a row -- is rewarded (+2), as is correctly skipping after asking (+3).
The initial utterance is the user answer to the start node question, subsequent utterances are given by (paraphrases of) answers leading to the next goal node.

To represent users with a concrete information goal (free mode), the simulator randomly selects a single, static goal and constraints for the entire dialog.
A random node (with at least one associated FAQ question) is chosen from the graph as the user goal and one of the FAQ questions associated with that node is chosen as the initial user utterance. 
The simulator then finds a valid path from the start node to the goal node, saving the user answers along that path.
Each time the dialog agent asks a question from a node along the goal path, the simulator can use the stored answers to respond.  
In case the node is not along the goal path, a random answer will be chosen to continue the dialog.
To discourage long dialogs, each turn is given a small negative reward (-1).
To discourage asking unnecessary questions, any questions that are not part of the stored goal path are punished (-4).
Finally, reaching the goal node and outputting its content is given a reward of $4\times tree\_depth$ (e.g. in the REIMBURSE dataset: reward of +128).

A dialog will stop after: 1) reaching 50 dialog turns, 2) presenting the user goal, or 3) presenting the same node 3 times (user patience).
To simulate unseen text input, noise can be added to user utterances in the form of a normal distribution around the original utterance encoding vector $\textbf{u}$, using a percentage $n$ of $u$ as standard deviation: $\mathcal{N}(\mathbf{u}, n |\textbf{u}|)$.

\subsection{Baseline}

\begin{figure}[t]
    \centering
    \includegraphics[width=0.45\textwidth]{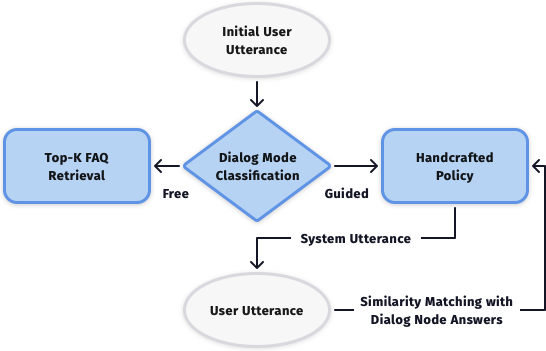}
    \caption{Baseline architecture: Combines an FAQ retrieval system (free mode) and a handcrafted dialog system (guided mode), with a classifier deciding which policy is active based on the input in the first turn.}
    \label{fig:baseline}
\end{figure}

For the baseline system (see Figure \ref{fig:baseline}), we combine an FAQ retrieval system (free mode) and a handcrafted dialog system (guided mode) together, training a classifier to decide which policy is active based on the input in the first turn.

To train the dialog mode classifier, we fine-tune a German BERT model \cite{gbert}, providing a user utterance and the associated node text as inputs (see Appendix \ref{appendix:train_details}).
In free mode, a state-of-the-art similarity model \cite{sentencetransformers} is used to compare the first user utterance to all nodes in the dialog tree, directly outputting the most similar match to the user.
In guided mode, node text is outputted to the user at each node. 
Their response is then compared to the prototypical answers for that node (using the same similarity model), and the policy then skips to the node connected by the most similar prototypical answer.

\subsection{Reinforcement Learning Model}

We propose a novel RL-architecture based on Dueling Network Architectures (DDQN) \cite{duelingDQN}. 
For improved stability and convergence speed, we integrate Munchausen Reinforcement Learning \cite{vieillard2020munchausen}, Double Q-Learning \cite{doubleDQN}, and Hindsight Experience Replay \cite{hindsightExperienceReplay} in conjunction with Loss-Adjusted Prioritized Experience Replay \cite{lapreplay}.

\begin{figure*}[htb]
    \centering
    \includegraphics[width=0.95\textwidth]{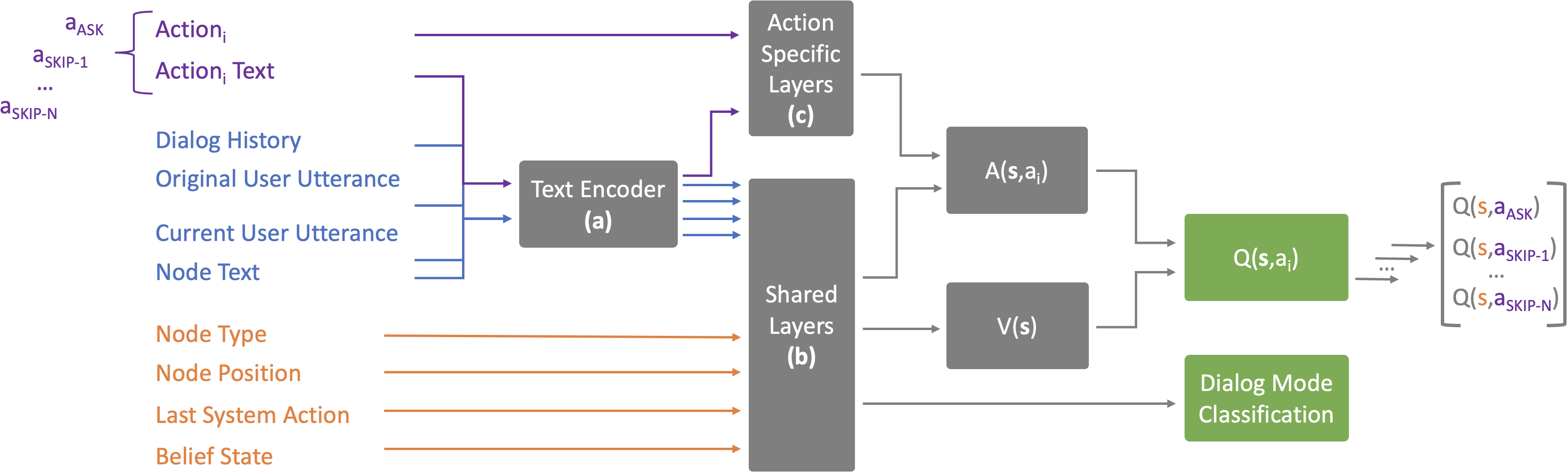}
    \caption{Proposed RL architecture: The state-value function $V(\mathbf{s})$ for state $\mathbf{s}$ is calculated from the shared layers (b), independent of action inputs. For the advantage function $A(\mathbf{s},a)$, the outputs from shared layers (b) and action-specific layers (c) are combined. The final state-action-values $Q$ are obtained by performing one forward pass per action $a_i$, each yielding a scalar $q_{\mathbf{s},a_i}$, which are then concatenated by state into a vector with one value per action. Additionally, a binary dialog mode classifier is added on top of the shared layers (b).} 
    \label{fig:architecture}
\end{figure*}

As a deviation from conventional DQN-based \cite{dqn} algorithms, we re-parameterize the usual network structure   
$\mathbf{Q}: \mathbf{s}, \pmb{\theta} \mapsto \mathbb{R}^{|A|}$
to
$Q: \mathbf{s}, a_i,  \pmb{\theta} \mapsto \mathbb{R}$, 
where $\mathbf{s}$ is a state vector, $\pmb{\theta}$ represents the trainable network weights, and $a_i$ is the $i$-th action in action space $A$ with $|A|$ discrete actions.
Thus, our architecture has one output node instead of $|A|$ output nodes (see Figure \ref{fig:architecture}).
Instead of performing one forward pass per state, we now perform $n(\mathbf{s})$ forward passes per state, where $n(\mathbf{s}) \leq |A|$ is the variable action count at state $\mathbf{s}$, concatenating all outputs to obtain the full state-action vector $\mathbf{Q}(\mathbf{s}, \pmb{\theta})$.
By batching these forward passes, we achieve comparable runtime performance.
The benefits of this approach are twofold: 
1) It scales to an arbitrary number of actions without increasing the number of output neurons.
2) We can process action-specific inputs, e.g., action text (Figure \ref{fig:architecture} (a)), allowing the model to infer information from text rather than just exploiting action-space structure.

To keep the state values stable across different actions in the same state, we expand on the idea of DDQN \cite{duelingDQN}, which processes inputs using shared layers (Figure \ref{fig:architecture} (b)) and calculates a state-value $V(\mathbf{s})$ and an advantage function $A(\mathbf{s},a)$ with separate network layers on top of the shared layer.
Here, we add separate layers for encoding action inputs (Figure \ref{fig:architecture} (c)), which we then concatenate with the shared layer output (Figure \ref{fig:architecture} (b)) in order to calculate $A(\mathbf{s},a)$.
$V(\mathbf{s})$ only receives inputs based on state $s$, and is thus decoupled from action-related inputs.
We found this stabilized Q-values and performed better than the original algorithms in all our experiments (e.g. DDQN: 31.87\% combined success on test data).

After experimenting with an external dialog mode classifier, we found that adding dialog mode classification as an auxiliary task improved our success metric.
Therefore, we add an additional binary classification head to the output of the shared state layers from our enhanced DDQN to predict which dialog mode the current turn belongs to.
Using cross entropy loss, we add this task as a hard-parameter sharing multitask learning objective \cite{caruana1993multitask} in order to force the model to develop an understanding of the different tasks: $\mathcal{L} = \mathcal{L}_{\text{ddqn}} + \lambda \mathcal{L}_{\text{intent}}$.

\section{REIMBURSE: A Dataset for CTS}
\label{sec:data}
For this task, we collect and publish a dataset from two sources: a dialog tree, defining the general system behavior, and a corpus of user utterances, with paraphrases for both user questions and responses to system questions.
This dataset represents challenging real-world data, both in the complex structure of the dialog tree for this domain, and the real-world nature of the user utterances.
Examples of user questions and user responses are provided in Appendix \ref{appendix:dataset}.

\paragraph{Dialog Tree}
We provide a dialog tree for travel reimbursement, created by subject-area-experts using a graphical dialog designer tool.
The dialog tree consists of four different types of nodes: 
1) Dialog Nodes, defining a system question and possible user answers. 
2) Variable Nodes, defining a system question and storing the user answer for later use in the beliefstate. 
3) Information Nodes, defining information that the system should share with the user without expecting a user response. 
And 4) Logic Nodes, defining dialog flow based on logical conditions evaluated against beliefstate values, without outputting anything to the user. 
An example of each of these can be seen in Figure \ref{fig:dialog_designer} in the Appendix.

In total, the dialog graph we use contains 123 total nodes (with a maximum depth of 32), of which, 79 are information nodes which contain answers to user questions and 23 are Dialog Nodes with system questions for the user.
The maximum action count per node (directly connected nodes) is 14.

\paragraph{User Utterances}

We collected a real world corpus, called REIMBURSE. This corpus consists of 452 free dialog questions, each corresponding to an Information Node in the dialog tree, with an average of 5.72 paraphrases per Information Node. 
This was done through recording user utterances in two early pilot studies, and manual augmentation by two expert annotators.
Synonyms for user responses to system questions were collected the same way, resulting in 408 total response paraphrases, on average 5.58 per user answer. For examples, see Appendix \ref{appendix:utterances}.

\section{Experimental Setup}

The RL agents receive the following inputs: beliefstate, last system action type, action index, and current node type encoded by one-hot vectors, current node position as a binary tree encoding \cite{treeencoding},
the dialog history, initial user utterance, current user utterance, current node text and the node's prototypical answer texts encoded by either an English-German RoBERTa \cite{liu2019roberta} \footnote{https://huggingface.co/T-Systems-onsite/cross-en-de-roberta-sentence-transformer?doi=true} or sentence transformers \cite{sentencetransformers} model.
The model is trained against the user simulator using the Adam optimizer \cite{kingma2014adam} with a learning rate of $1e^{-4}$. 
We set the maximum number of training dialog turns to $1.5$M and save the model which performs  best in evaluation (performed every $10$k turns with $500$ simulated dialogs) for testing. 
Evaluation and testing are also performed against the user simulator. 
However, in this mode agent exploration is disabled. For more details on model and training parameters, see Appendix \ref{appendix:train_details}.

Testing is done on 500 simulated dialogs (in RQ3 generated from held out utterances).
We measure success and skip ratios for guided and free mode separately and jointly.
To check task understanding, we measure the dialog mode prediction consistency: 
we calculate the difference between the percentage of turns predicted for guided and free mode per dialog, then averaging across all dialogs.

\subsection{RQ1: Task Performance}

To understand how well the RL agent learns to support both dialog modes, we measure combined task success as well as success on each individual mode (how often user reaches goal), the mode-specific skip ratios (higher skip ratio desired for free, lower for guided mode), and the mode prediction and consistency measures (proxy for task understanding).
All models are trained and tested against the user simulator generating dialogs from the same distribution, following standard RL procedure. We apply a noise level of $n=10\%$ to encourage generalization.

\subsection{RQ2: Task Performance in a Noisy Environment}

To test how well our models perform in noisy environments, we test the success and mode classification F1 score of models trained in RQ1 on simulated dialogs with increasing noise levels.
We test five times, reporting average results and statistically significant performance differences.

\subsection{RQ3: Generalizing to New Data}
In the previous experiments, we have followed the RL convention of testing our models on new dialogs generated from the same user simulator they trained on, with the same set of possible user utterances.
However, much of the challenge in this domain comes from understanding text input.
By exchanging the set of user utterances available to the simulator for testing, we explore how well the model can generalize to user inputs not seen during training.
To this end, we split our corpus into a test/eval and a train set. 
The train/eval set contains 279 FAQ questions (3.5 questions per Information node) and 246 responses (3.4 paraphrases per response prototype).
The test set contains 173 user questions (2.2 questions per Information node) and 162 responses (2.2 paraphrases per response prototype).
We then train and evaluate the best performing models from RQ1 and RQ2 on simulated dialogs generated from the train split. 
Finally, to measure how well they can generalize, we test the final performance on dialogs generated from the unseen paraphrases in the test split.

\section{Results and Discussion}

\subsection{RQ1: Task Performance}
\begin{table*}[h]
    \center
    \resizebox{\textwidth}{!}{
         \begin{tabular}{ |c|c|c|c|c||c|c|c| } 
         \hline
          \textbf{Model} & 
          \makecell{\textbf{Success} \\ \textbf{(guided)}}  & 
          \makecell{\textbf{Skip Ratio} \\ \textbf{(guided)}} & 
          \makecell{\textbf{Success} \\ \textbf{(free)}} & 
          \makecell{\textbf{Skip Ratio} \\ \textbf{(free)}} & 
          \makecell{\textbf{Success} \\ \textbf{(combined)}} & 
          \makecell{\textbf{Dialog Mode} \\ \textbf{Prediction F1}} &
          \makecell{\textbf{Dialog Mode} \\ \textbf{Prediction Consistency}}
          \\ \hline

         Baseline        & 64.71\% & \textbf{0.0}  & 22.84\% & \textbf{1.0}  & 43.10\% & 0.83 & \textbf{1.0 }\\
         RoBERTa CTS     & 75.12\% & 0.10 & \textbf{74.20}\% & 0.55 & 74.40\% &\textbf{ 1.0 }& \textbf{1.0} \\
         Sentencesim CTS & \textbf{84.77}\% & 0.07  & 72.00\% & 0.58 & \textbf{77.16}\% & \textbf{1.0} & \textbf{1.0}  \\
         \hline
        \end{tabular}
    }
    \caption{Average model performance on data with 10\% noise. Both RL agents are able to significantly outperform the baseline (t-test: $p < \num{3e-8}$) in terms of success (guided, free, and combined). The RL agents learned to distinguish between dialog modes, skipping more than half of nodes in free mode and less than 10\% in guided.}
    \label{tab:joint_data}
\end{table*}

To verify the user simulator, we evaluate it against our baseline model, without noise and only using the prototypical user answers from the dialog tree.
This yields a success rate of $99.35$\% for the guided task and an F1-score of $1.0$ for the dialog mode prediction, showing that the simulator works as expected. On the free task, it reaches $64.46$\% success. Given the challenging nature of the dataset, it is to be excepted that top-1 retrieval is not perfect.

Just by adding user answer paraphrases and $10$\% noise, baseline performance drops substantially on all tasks (see Table \ref{tab:joint_data}), demonstrating the task difficulty w.r.t. a real-world scenario.
Our RL agents significantly outperform the baseline (t-test $p<\num{3e-8}$ ) on all task success metrics, except the skip ratios (see Table \ref{tab:joint_data}).
This shows our models are better able to learn the task compared to the baseline, especially  under noisy text inputs, which we attribute to the improved generalization.

\paragraph{RQ1.1: Differentiating Between Dialog Modes}

Table \ref{tab:joint_data} shows that both of our models learn to skip more frequently in free mode (e.g., $0.58$) and less frequently in guided mode (e.g., $0.07$).
This indicates that multitask learning helped understand the user's intended mode, as we would otherwise not see a difference in skipping behaviour.
Additionally, the classification consistency is $1.0$, indicating stable task understanding.
Jointly learning intent prediction also improves dialog mode classification compared to using a pre-trained classifier.

\paragraph{RQ1.2: Asking Only Necessary Questions}

While the baseline has, by construction, perfect skip ratios on both free (1-step retrieval) and guided tasks (no skipping), the skip ratios in Table \ref{tab:joint_data} demonstrate that our models learn to ask questions in both modes. 
In free mode, we see that asking some questions helps differentiate the user goal, as we obtain much better task success than the 1-step baseline. High skip ratios (e.g., $0.58$) in this mode show the model still skips unnecessary nodes.

\subsection{RQ2: Task Performance in a Noisy Environment}

\begin{figure}[htb]
    \centering
    \includegraphics[width=0.45\textwidth]{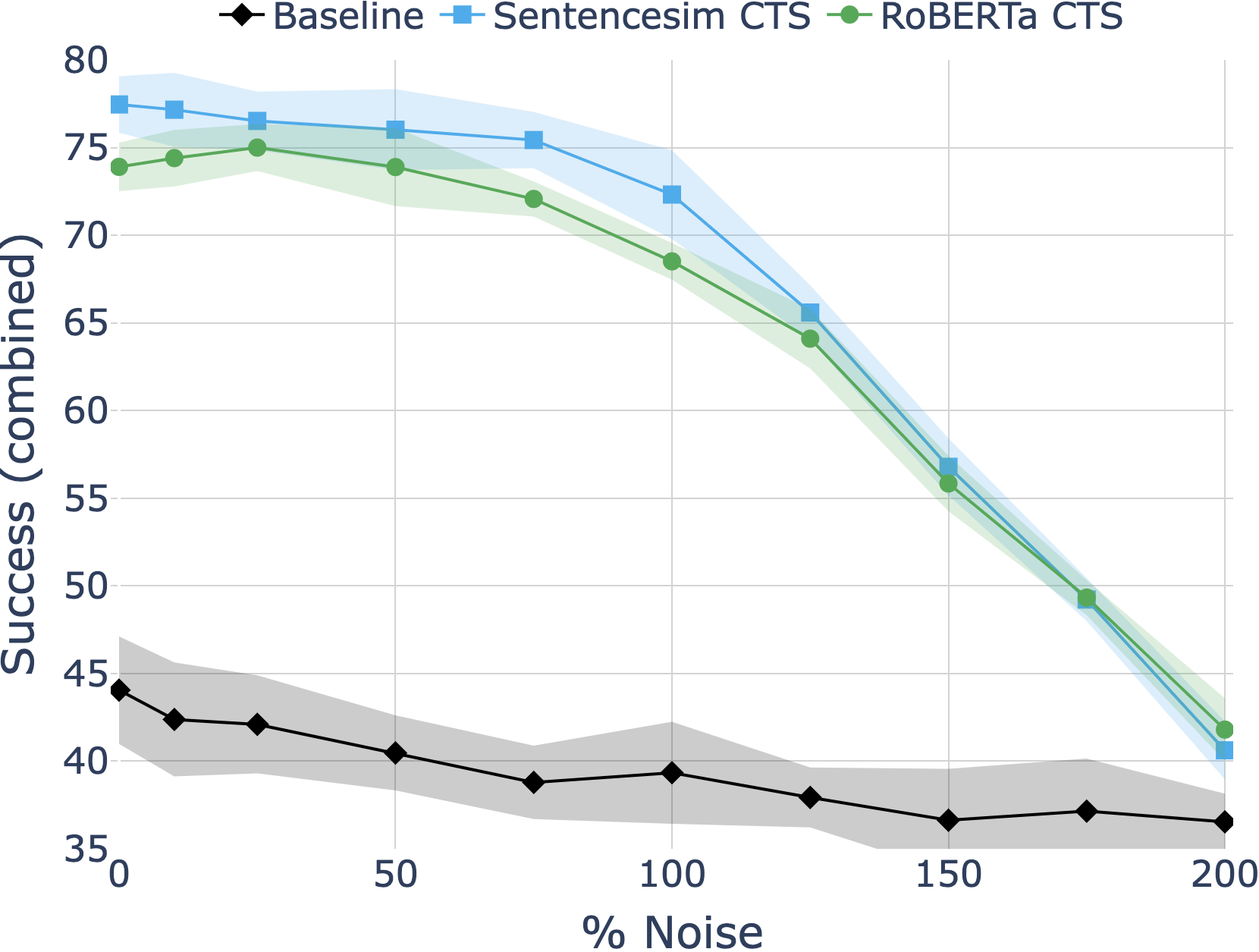}
    \caption{Task performance for different noise levels on user input (drawn from a normal distribution around the original text encoding vector $\textbf{u}$, using a percentage $n$ of $\mathbf{u}$ as standard deviation: $\mathcal{N}(\mathbf{u}, n |\textbf{u}|)$.
    }
    \label{fig:noise_graph}
\end{figure}

Figure \ref{fig:noise_graph} shows our models are able to handle high levels of noise: performance only decreases rapidly after 100\% noise.
Our models demonstrate a significant performance increase over the baseline at all noise levels (t-test $p<0.003$).
Additionally, we find that dialog mode understanding is robust w.r.t. input noise (dialog mode classification F1 stays unchanged at $1.0$ after rounding, e.g. consistency only drops from $1.0$ without noise to $0.83$ at the highest noise setting).
The robustness to noise suggests that the regularization techniques are effective.

\subsection{RQ3: Text Understanding and Generalization to New Data}

\begin{table*}[h!]
    \center
    \resizebox{\textwidth}{!}{
        \begin{tabular}{ |l|c|c|c|c||c|c|c| } 
         \hline
          \textbf{Model} & 
          \makecell{\textbf{Success} \\ \textbf{(guided)}}  & 
          \makecell{\textbf{Skip Ratio} \\ \textbf{(guided)}} & 
          \makecell{\textbf{Success} \\ \textbf{(free)}} & 
          \makecell{\textbf{Skip Ratio} \\ \textbf{(free)}} & 
          \makecell{\textbf{Success} \\ \textbf{(combined)}} & 
          \makecell{\textbf{Dialog Mode} \\ \textbf{Prediction F1}} &
          \makecell{\textbf{Dialog Mode} \\ \textbf{Prediction Consistency}}
          \\ \hline

         \multicolumn{8}{|c|}{\textbf{Evaluation Data}}  \\ \hline
         Baseline        & 81.05\% & \textbf{0.0}  & 25.54\% & \textbf{1.0}  & 55.40\% & 0.95 & \textbf{1.0} \\
         Sentencesim CTS & 82.96\% & 0.13  & 71.19\% & 0.54 & 76.01\% & \textbf{1.0} & \textbf{1.0 }\\
         RoBERTa CTS     & \textbf{85.36}\% & 0.07 & \textbf{86.18}\% & 0.53 & \textbf{85.81}\% & \textbf{1.0} & \textbf{1.0 }\\

         \hline
         \multicolumn{8}{|c|}{\textbf{Test Data}}  \\
         \hline
         Baseline        & 65.52\% & \textbf{0.0}  & 20.38\% & \textbf{1.0}  & 42.05\% & 0.85 & \textbf{1.0} \\
         Sentencesim CTS & \textbf{74.76}\% & 0.19 & 54.38\% & 0.61 & 62.54\% & \textbf{0.88} & 0.96 \\
         \textbf{RoBERTa CTS}     & 57.10\% & 0.25  & 66.78\% & 0.53 & \textbf{62.58}\% & 0.85 & \textbf{1.0} \\
         - Action Positions    & 46.70 \% & 0.29 & \textbf{66.79}\% & 0.55 & 57.59\% & 0.82 & 0.99 \\
         - Action Text    & 57.63\% & 0.23 & 55.31\% & 0.57 & 55.28\% & 0.84 & 0.98 \\
         - Node Text     & 39.82\% & 0.20 & 55.34\% & 0.54 & 47.95\% & 0.83 & 0.96 \\
         - Node Positions    & 54.98\% & 0.20 & 55.40\% & 0.59 & 55.21\% & 0.81 & 0.99 \\
         - Node Type   & 50.67\% & 0.20 & 64.41\% & 0.56 & 58.39\% & 0.84 & 0.98 \\
         - Mode Prediction     & 43.61\% & 0.23 & 60.38\% & 0.59 & 52.33\% & n/a & n/a \\
         - Beliefstate    & 62.26\% & 0.21 & 42.66\% & 0.57 & 51.05\% & 0.82 & 0.99 \\
         \hline
        \end{tabular}
    }
    \caption{Model performance on simulated dialogs generated from dataset splits, and skip ratios (skipped dialog nodes w.r.t. dialog length). Here, higher skip ratios are better for the free setting (looking for a direct answer), but lower ones are preferable for guided mode (exploratory dialogs).
    Rows underneath \emph{RoBERTa CTS} for the data represent input ablations, i.e. the listed input was removed from the model.}
    \label{tab:automatic_results}
\end{table*}

When only training on a fraction of the data and testing on the unseen split, we observe some performance drops for the baseline and all RL agents, compared to the setting from RQ1, as well as between evaluation and test performance in RQ3 (see Table \ref{tab:automatic_results}).
The performance drop in the RL agents can likely be explained by the drop in the dialog mode classification score (e.g., from $1.0$ in RQ1 to $0.85$).
This paired with high dialog mode prediction consistency ($0.96$, $1.0$) means the models more often classified the dialog mode incorrectly and could not recover during a dialog.
This misclassification likely also explains increased skipping in guided mode (e.g., from $0.07$ in RQ1 to $0.19$).

However, even in this more challenging setting, our models demonstrate much better combined task success than the baseline. 
This is particularly interesting as utterances in the test data were never seen in training, meaning the model could not have merely exploited structural properties of the data, but rather must have learned text understanding to solve the task. 
For example (see Appendix \ref{appendix:dialogs:free:directskip}), the model can understand that when a user asks what to do after an earthquake, they need the emergency help number and skips directly to that node, even though nothing about earthquakes appeared in training.
The system also learns to recognize when information is missing from a user question and ask for additional information, e.g., if the business trip took place in or outside the home city, even for new questions (see Appendix \ref{appendix:dialogs:free:clarifying}).  

To better understand how the model processed inputs, we performed an ablation study (Table \ref{tab:automatic_results}) of the RoBERTa agent.
From the success rates without either action text or action position, we see that providing action context in input space increases model performance, validating our new architecture.

In free mode, when node or action text are removed, there is a large drop in model performance (about 10\%), which is not as severe if instead action positions are left out.
This suggests that the model learns to understand text as it can correlate user utterances to text from nodes and actions in the dialog graph, even with noise and  synonyms not seen during training. 
Node position intuitively simplifies path finding, which explains a drop in performance when leaving it out.
It also makes sense then that removing the dialog history has a similar effect, as it encodes path information in textual form.
We conclude from this that both textual and positional information are needed for this task and better understanding of either input could likely boost performance.

In the guided mode, we observe that while node text was very important (nearly 20\% drop), removing action text had almost no effect.
This might relate to the text length associated with each action, and the length of user input per node (usually much less information than in the free mode questions).
As user input was often one to three words long, it could have been easier for the model to learn to correlate the user input to an action position rather than action text, especially given that BERT-based language models rely on context.
Thus, improving focus on action text in the guided mode is an area for future investigation.
Removing the node position or dialog history had a minor impact, which agrees well with the objective of the guided mode: it should not rely much on path planning, instead user input should be considered at most nodes to move the dialog along.

In general, dialog mode prediction improves performance in both dialog modes.
Without this, the agent is able to learn to skip more in free mode and less in guided, but success drops.
In the multitask learning setting, we force the model to learn to output a consistent prediction across a dialog (Table \ref{tab:automatic_results} consistency values are close to $1.0$ for all multitask models), which may explain the higher success. 

\section{Related Work}

\paragraph{Data Efficient Task Oriented Dialog Systems}
In some domains, e.g.,  recommendation, the number of actions for a dialog system can be quite large prompting new architectures which group the actions into a tree structure to reduce the search space \cite{chen2019large, actor-critique}.
While the information seeking dialogs have different characteristics, including a tree structure for the data, can help experts maintain control of the system and reduce the action search space.


\paragraph{Task Oriented Systems with Expert involvement}
In many domains, it is crucial domain experts maintain control of dialog flow to ensure correctness of system outputs. 
An early approach \cite{williams2008integrating} involved a hand-crafted dialog policy which  output the allowed actions at a time step and a POMDP policy then chose the optimal action from this set.
More recent approaches, e.g., hybrid code networks (HCN) have expanded on this idea for neural systems \cite{williams-etal-2017-hybrid, liang2018hierarchical, razumovskaia2019incorporating}, where action space can be constrained using masks.
\citet{shukla2020conversation} extend the HCN approach by automatically converting expert designed dialog trees into hybrid code networks (HCN).
This approach increases explainability of system behavior, but doesn't provide a mechanism for skipping portions of a dialog irrelevant to a user, falling short of addressing free dialog.

\paragraph{Knowledge Augmented Dialog Systems}
The DSTC10 challenge \cite{kim2022knowledge} introduced a new track for knowledge grounded dialog, combining task-oriented dialog 
and FAQ style questions answered from unstructured documents. 
Such hybrid dialog systems help answer free form user queries, but still require a concrete question, and do not allow precise control over generated answers.

Another recent approach \cite{raghu-etal-2021-end} proposed grounding dialog systems on troubleshooting flow charts and FAQ questions. 
Here, the dialog system fuses knowledge from FAQs with an under-specified flow chart to generate an appropriate response to a user input. 
However, dialogs appear limited to  short flow charts, and response generation is again outside the designer's control. 

\paragraph{Clarifying Questions and Conversational Search} 
Conversational Search and clarifying questions are two intertwined areas of research which focus on disambguating free form user queries.
The goal of these tasks is to generate one or more questions to narrow down a user's information need \cite{zamani2020mimics} in order to retrieve relevant documents. 
To improve conversational search performance, much work has focused on ranking candidate questions \cite{kumar2020ranking} and leveraging the user answers \cite{bi2021asking}.
However, neither of these tasks do not allow for human oversight of the questions that will be asked, and currently do not consider any structure for the organization of questions.



\section{Conclusion and Future Work}

In this work, we introduced a new task, CTS which combines the positive aspects of both an FAQ retrieval and task-based dialog system, along with a real-world German language dataset, REIMBURSE, a domain-agnostic user simulator, and baseline system based on state-of-the-art language models.
Furthermore, we demonstrated a novel, scalable RL architecture, showing significant performance improvements over the baseline. Our models are able to handle high levels of noise in the input data and demonstrate the ability to learn text-based information from user utterances, as seen by their ability to handle unseen data.
In the future, it would be interesting to explore data augmentation for generating unseen utterances as well as to explore methods, particularly for guided mode dialogs, to encourage the model to rely even more information from text. 
We also hope that the community helps in exploring performance on additional languages and task instances.

\section{Limitations}
We recognize the following limitations in our work:
\begin{itemize}
    \item When testing on unseen real world data rather than the noisy simulation, we notice a performance drop between the evaluation and test results. As we are working in a very low resource setting and using a challenging real-world dataset, it is difficult to state with certainty if this is a result of the challenge inherent to the dataset or a weakness of the models to generalize.
    \item We were not able to perform testing in an interactive user study, but rather tested against simulated dialogs generated from pre-collected real-world utterances. While these were collected from real users of the dialog system, a user study would be valuable to investigate how the system handles input in a live setting.
    \item As we only collected a single dataset for the new CTS task, we have not investigated domain generalizability.
\end{itemize}

\bibliographystyle{acl_natbib}
\bibliography{aaai23}

\begin{thebibliography}{30}
\expandafter\ifx\csname natexlab\endcsname\relax\def\natexlab#1{#1}\fi

\bibitem[{Andrychowicz et~al.(2017)Andrychowicz, Crow, Ray, Schneider, Fong,
  Welinder, McGrew, Tobin, Abbeel, and Zaremba}]{hindsightExperienceReplay}
Marcin Andrychowicz, Dwight Crow, Alex Ray, Jonas Schneider, Rachel Fong, Peter
  Welinder, Bob McGrew, Josh Tobin, Pieter Abbeel, and Wojciech Zaremba. 2017.
\newblock \href
  {https://proceedings.neurips.cc/paper/2017/hash/453fadbd8a1a3af50a9df4df899537b5-Abstract.html}
  {Hindsight experience replay}.
\newblock pages 5048--5058.

\bibitem[{Bi et~al.(2021)Bi, Ai, and Croft}]{bi2021asking}
Keping Bi, Qingyao Ai, and W.~Bruce Croft. 2021.
\newblock \href {https://doi.org/10.1145/3471158.3472232} {Asking clarifying
  questions based on negative feedback in conversational search}.
\newblock In \emph{{ICTIR} '21: The 2021 {ACM} {SIGIR} International Conference
  on the Theory of Information Retrieval, Virtual Event, Canada, July 11,
  2021}, pages 157--166. {ACM}.

\bibitem[{Caruana(1993)}]{caruana1993multitask}
Rich Caruana. 1993.
\newblock \href {https://doi.org/10.1016/b978-1-55860-307-3.50012-5} {Multitask
  learning: {A} knowledge-based source of inductive bias}.
\newblock In \emph{Machine Learning, Proceedings of the Tenth International
  Conference, University of Massachusetts, Amherst, MA, USA, June 27-29, 1993},
  pages 41--48. Morgan Kaufmann.

\bibitem[{Chan et~al.(2020)Chan, Schweter, and M{\"o}ller}]{gbert}
Branden Chan, Stefan Schweter, and Timo M{\"o}ller. 2020.
\newblock \href {https://doi.org/10.18653/v1/2020.coling-main.598}
  {{G}erman{'}s next language model}.
\newblock In \emph{Proceedings of the 28th International Conference on
  Computational Linguistics}, pages 6788--6796, Barcelona, Spain (Online).
  International Committee on Computational Linguistics.

\bibitem[{Chen et~al.(2019)Chen, Dai, Cai, Zhang, Wang, Tang, Zhang, and
  Yu}]{chen2019large}
Haokun Chen, Xinyi Dai, Han Cai, Weinan Zhang, Xuejian Wang, Ruiming Tang,
  Yuzhou Zhang, and Yong Yu. 2019.
\newblock \href {https://doi.org/10.1609/aaai.v33i01.33013312} {Large-scale
  interactive recommendation with tree-structured policy gradient}.
\newblock In \emph{The Thirty-Third {AAAI} Conference on Artificial
  Intelligence, {AAAI} 2019, The Thirty-First Innovative Applications of
  Artificial Intelligence Conference, {IAAI} 2019, The Ninth {AAAI} Symposium
  on Educational Advances in Artificial Intelligence, {EAAI} 2019, Honolulu,
  Hawaii, USA, January 27 - February 1, 2019}, pages 3312--3320. {AAAI} Press.

\bibitem[{Cohen(2020)}]{cohen2020back}
Philip~R. Cohen. 2020.
\newblock \href {https://ojs.aaai.org/index.php/AAAI/article/view/7073} {Back
  to the future for dialogue research}.
\newblock In \emph{The Thirty-Fourth {AAAI} Conference on Artificial
  Intelligence, {AAAI} 2020, The Thirty-Second Innovative Applications of
  Artificial Intelligence Conference, {IAAI} 2020, The Tenth {AAAI} Symposium
  on Educational Advances in Artificial Intelligence, {EAAI} 2020, New York,
  NY, USA, February 7-12, 2020}, pages 13514--13519. {AAAI} Press.

\bibitem[{Fujimoto et~al.(2020)Fujimoto, Meger, and Precup}]{lapreplay}
Scott Fujimoto, David Meger, and Doina Precup. 2020.
\newblock \href
  {https://proceedings.neurips.cc/paper/2020/hash/a3bf6e4db673b6449c2f7d13ee6ec9c0-Abstract.html}
  {An equivalence between loss functions and non-uniform sampling in experience
  replay}.

\bibitem[{Gao et~al.(2018)Gao, Galley, and Li}]{10.1145/3209978.3210183}
Jianfeng Gao, Michel Galley, and Lihong Li. 2018.
\newblock \href {https://doi.org/10.1145/3209978.3210183} {Neural approaches to
  conversational ai}.
\newblock In \emph{The 41st International ACM SIGIR Conference on Research \&
  Development in Information Retrieval}, SIGIR '18, page 1371–1374, New York,
  NY, USA. Association for Computing Machinery.

\bibitem[{Kim et~al.(2022)Kim, Liu, Jin, Papangelis, Hedayatnia,
  Gopalakrishnan, and Hakkani-T{\"u}r}]{kim2022knowledge}
Seokhwan Kim, Yang Liu, Di~Jin, Alexandros Papangelis, Behnam Hedayatnia,
  Karthik Gopalakrishnan, and Dilek Hakkani-T{\"u}r. 2022.
\newblock Knowledge-grounded task-oriented dialogue modeling on spoken
  conversations track at dstc10.

\bibitem[{Kingma and Ba(2015)}]{kingma2014adam}
Diederik~P. Kingma and Jimmy Ba. 2015.
\newblock \href {http://arxiv.org/abs/1412.6980} {Adam: {A} method for
  stochastic optimization}.

\bibitem[{Kumar et~al.(2020)Kumar, Raunak, and Callan}]{kumar2020ranking}
Vaibhav Kumar, Vikas Raunak, and Jamie Callan. 2020.
\newblock \href {https://doi.org/10.1145/3340531.3412137} {Ranking
  clarification questions via natural language inference}.
\newblock In \emph{{CIKM} '20: The 29th {ACM} International Conference on
  Information and Knowledge Management, Virtual Event, Ireland, October 19-23,
  2020}, pages 2093--2096. {ACM}.

\bibitem[{Liang and Yang(2018)}]{liang2018hierarchical}
Weiri Liang and Meng Yang. 2018.
\newblock \href {https://doi.org/10.1007/978-3-319-95933-7\_24} {Hierarchical
  hybrid code networks for task-oriented dialogue}.
\newblock In \emph{Intelligent Computing Theories and Application - 14th
  International Conference, {ICIC} 2018, Wuhan, China, August 15-18, 2018,
  Proceedings, Part {II}}, volume 10955 of \emph{Lecture Notes in Computer
  Science}, pages 194--204. Springer.

\bibitem[{Liu et~al.(2019)Liu, Ott, Goyal, Du, Joshi, Chen, Levy, Lewis,
  Zettlemoyer, and Stoyanov}]{liu2019roberta}
Yinhan Liu, Myle Ott, Naman Goyal, Jingfei Du, Mandar Joshi, Danqi Chen, Omer
  Levy, Mike Lewis, Luke Zettlemoyer, and Veselin Stoyanov. 2019.
\newblock \href {http://arxiv.org/abs/1907.11692} {Roberta: {A} robustly
  optimized {BERT} pretraining approach}.
\newblock \emph{CoRR}, abs/1907.11692.

\bibitem[{Mnih et~al.(2013)Mnih, Kavukcuoglu, Silver, Graves, Antonoglou,
  Wierstra, and Riedmiller}]{dqn}
Volodymyr Mnih, Koray Kavukcuoglu, David Silver, Alex Graves, Ioannis
  Antonoglou, Daan Wierstra, and Martin~A. Riedmiller. 2013.
\newblock \href {http://arxiv.org/abs/1312.5602} {Playing atari with deep
  reinforcement learning}.
\newblock volume abs/1312.5602.

\bibitem[{Montazeralghaem et~al.(2021)Montazeralghaem, Allan, and
  Thomas}]{actor-critique}
Ali Montazeralghaem, James Allan, and Philip~S. Thomas. 2021.
\newblock \href {https://doi.org/10.1145/3460231.3474271} {Large-scale
  interactive conversational recommendation system using actor-critic
  framework}.
\newblock In \emph{Fifteenth ACM Conference on Recommender Systems}, RecSys
  '21, page 220–229, New York, NY, USA. Association for Computing Machinery.

\bibitem[{Raghu et~al.(2021)Raghu, Agarwal, Joshi, and
  {Mausam}}]{raghu-etal-2021-end}
Dinesh Raghu, Shantanu Agarwal, Sachindra Joshi, and {Mausam}. 2021.
\newblock \href {https://doi.org/10.18653/v1/2021.emnlp-main.357} {End-to-end
  learning of flowchart grounded task-oriented dialogs}.
\newblock In \emph{Proceedings of the 2021 Conference on Empirical Methods in
  Natural Language Processing}, pages 4348--4366, Online and Punta Cana,
  Dominican Republic. Association for Computational Linguistics.

\bibitem[{Razumovskaia and Eskenazi(2019)}]{razumovskaia2019incorporating}
Evgeniia Razumovskaia and Maxine Eskenazi. 2019.
\newblock Incorporating rules into end-to-end dialog systems.
\newblock In \emph{Proc. 3rd NeurIPS Workshop on Conversational AI, Vancouver,
  Canada}, pages 1--11.

\bibitem[{Reimers and Gurevych(2019)}]{sentencetransformers}
Nils Reimers and Iryna Gurevych. 2019.
\newblock \href {https://doi.org/10.18653/v1/D19-1410} {Sentence-bert: Sentence
  embeddings using siamese bert-networks}.
\newblock In \emph{Proceedings of the 2019 Conference on Empirical Methods in
  Natural Language Processing and the 9th International Joint Conference on
  Natural Language Processing, {EMNLP-IJCNLP} 2019, Hong Kong, China, November
  3-7, 2019}, pages 3980--3990. Association for Computational Linguistics.

\bibitem[{Shiv and Quirk(2019)}]{treeencoding}
Vighnesh Shiv and Chris Quirk. 2019.
\newblock \href
  {https://proceedings.neurips.cc/paper/2019/file/6e0917469214d8fbd8c517dcdc6b8dcf-Paper.pdf}
  {Novel positional encodings to enable tree-based transformers}.
\newblock In \emph{Advances in Neural Information Processing Systems},
  volume~32. Curran Associates, Inc.

\bibitem[{Shukla et~al.(2020)Shukla, Liden, Shayandeh, Kamal, Li, Mazzola,
  Park, Peng, and Gao}]{shukla2020conversation}
Swadheen Shukla, Lars Liden, Shahin Shayandeh, Eslam Kamal, Jinchao Li, Matt
  Mazzola, Thomas Park, Baolin Peng, and Jianfeng Gao. 2020.
\newblock \href {https://doi.org/10.18653/v1/2020.acl-demos.39} {{C}onversation
  {L}earner - a machine teaching tool for building dialog managers for
  task-oriented dialog systems}.
\newblock In \emph{Proceedings of the 58th Annual Meeting of the Association
  for Computational Linguistics: System Demonstrations}, pages 343--349,
  Online. Association for Computational Linguistics.

\bibitem[{Thakur et~al.(2021)Thakur, Reimers, R{\"u}ckl{\'e}, Srivastava, and
  Gurevych}]{thakur2021beir}
Nandan Thakur, Nils Reimers, Andreas R{\"u}ckl{\'e}, Abhishek Srivastava, and
  Iryna Gurevych. 2021.
\newblock \href
  {https://datasets-benchmarks-proceedings.neurips.cc/paper/2021/file/65b9eea6e1cc6bb9f0cd2a47751a186f-Paper-round2.pdf}
  {Beir: A heterogenous benchmark for zero-shot evaluation of information
  retrieval models}.
\newblock In \emph{35th Conference on Neural Information Processing Systems
  (NeurIPS 2021) Track on Datasets and Benchmarks}.

\bibitem[{van Hasselt et~al.(2016)van Hasselt, Guez, and Silver}]{doubleDQN}
Hado van Hasselt, Arthur Guez, and David Silver. 2016.
\newblock \href
  {http://www.aaai.org/ocs/index.php/AAAI/AAAI16/paper/view/12389} {Deep
  reinforcement learning with double q-learning}.
\newblock In \emph{Proceedings of the Thirtieth {AAAI} Conference on Artificial
  Intelligence, February 12-17, 2016, Phoenix, Arizona, {USA}}, pages
  2094--2100. {AAAI} Press.

\bibitem[{Vieillard et~al.(2020)Vieillard, Pietquin, and
  Geist}]{vieillard2020munchausen}
Nino Vieillard, Olivier Pietquin, and Matthieu Geist. 2020.
\newblock \href
  {https://proceedings.neurips.cc/paper/2020/hash/2c6a0bae0f071cbbf0bb3d5b11d90a82-Abstract.html}
  {Munchausen reinforcement learning}.

\bibitem[{Wang et~al.(2016)Wang, Schaul, Hessel, van Hasselt, Lanctot, and
  de~Freitas}]{duelingDQN}
Ziyu Wang, Tom Schaul, Matteo Hessel, Hado van Hasselt, Marc Lanctot, and Nando
  de~Freitas. 2016.
\newblock \href {http://proceedings.mlr.press/v48/wangf16.html} {Dueling
  network architectures for deep reinforcement learning}.
\newblock In \emph{Proceedings of the 33nd International Conference on Machine
  Learning, {ICML} 2016, New York City, NY, USA, June 19-24, 2016}, volume~48
  of \emph{{JMLR} Workshop and Conference Proceedings}, pages 1995--2003.
  JMLR.org.

\bibitem[{Williams(2008)}]{williams2008integrating}
Jason~D Williams. 2008.
\newblock Integrating expert knowledge into pomdp optimization for spoken
  dialog systems.
\newblock In \emph{Proceedings of the AAAI-08 Workshop on Advancements in POMDP
  Solvers}, volume~2, page~25.

\bibitem[{Williams et~al.(2017)Williams, Asadi, and
  Zweig}]{williams-etal-2017-hybrid}
Jason~D. Williams, Kavosh Asadi, and Geoffrey Zweig. 2017.
\newblock \href {https://doi.org/10.18653/v1/P17-1062} {Hybrid code networks:
  practical and efficient end-to-end dialog control with supervised and
  reinforcement learning}.
\newblock In \emph{Proceedings of the 55th Annual Meeting of the Association
  for Computational Linguistics (Volume 1: Long Papers)}, pages 665--677,
  Vancouver, Canada. Association for Computational Linguistics.

\bibitem[{Wolf et~al.(2020)Wolf, Debut, Sanh, Chaumond, Delangue, Moi, Cistac,
  Rault, Louf, Funtowicz, Davison, Shleifer, von Platen, Ma, Jernite, Plu, Xu,
  Scao, Gugger, Drame, Lhoest, and Rush}]{huggingface}
Thomas Wolf, Lysandre Debut, Victor Sanh, Julien Chaumond, Clement Delangue,
  Anthony Moi, Pierric Cistac, Tim Rault, Rémi Louf, Morgan Funtowicz, Joe
  Davison, Sam Shleifer, Patrick von Platen, Clara Ma, Yacine Jernite, Julien
  Plu, Canwen Xu, Teven~Le Scao, Sylvain Gugger, Mariama Drame, Quentin Lhoest,
  and Alexander~M. Rush. 2020.
\newblock \href {https://www.aclweb.org/anthology/2020.emnlp-demos.6}
  {Transformers: State-of-the-art natural language processing}.
\newblock In \emph{Proceedings of the 2020 Conference on Empirical Methods in
  Natural Language Processing: System Demonstrations}, pages 38--45, Online.
  Association for Computational Linguistics.

\bibitem[{Wu et~al.(2005)Wu, Yeh, and Chen}]{10.1145/1066078.1066079}
Chung-Hsien Wu, Jui-Feng Yeh, and Ming-Jun Chen. 2005.
\newblock \href {https://doi.org/10.1145/1066078.1066079} {Domain-specific faq
  retrieval using independent aspects}.
\newblock \emph{ACM Transactions on Asian Language Information Processing},
  4(1):1–17.

\bibitem[{Zamani et~al.(2020)Zamani, Lueck, Chen, Quispe, Luu, and
  Craswell}]{zamani2020mimics}
Hamed Zamani, Gord Lueck, Everest Chen, Rodolfo Quispe, Flint Luu, and Nick
  Craswell. 2020.
\newblock \href {https://doi.org/10.1145/3340531.3412772} {{MIMICS:} {A}
  large-scale data collection for search clarification}.
\newblock In \emph{{CIKM} '20: The 29th {ACM} International Conference on
  Information and Knowledge Management, Virtual Event, Ireland, October 19-23,
  2020}, pages 3189--3196. {ACM}.

\bibitem[{Zhang et~al.(2020)Zhang, Takanobu, Zhu, Huang, and
  Zhu}]{zhang2020recent}
Zheng Zhang, Ryuichi Takanobu, Qi~Zhu, MinLie Huang, and XiaoYan Zhu. 2020.
\newblock \href {https://doi.org/10.1007/s11431-020-1692-3} {Recent advances
  and challenges in task-oriented dialog systems}.
\newblock \emph{Science China Technological Sciences}, 63(10):2011--2027.

\end{thebibliography}

\appendix
\onecolumn

\section{User Simulator}
\label{appendix:simulator}
The (simplified) behavior of the user simulator is show below. To generate dialogs based on unseen utterances, one can simply replace the files containing questions and answers with new files containing different questions and answers than were present during the training process.

\begin{algorithm}
\begin{algorithmic}
\Require $\text{questions} \gets \text{List of Questions associated with dialog tree nodes}$
\Require $\text{answers} \gets \text{List of Answer synonyms}$
\Require $N \gets \text{Number of dialogs to be simulated}$
\Require $G(V,E)$ \Comment{Dialog Tree}
\Require $T \gets \text{Maximum number of turns per dialog}$

\State $n \gets 0$
\For{$n < N$}

    \State $v \gets \text{start node}$
    \State $m \gets \text{randomly choose GUIDED or FREE}$
    
    \If{$ m = \text{GUIDED}$} 
        \State $g \gets \text{randomly choose neighbour(v)}$ \Comment{Draw new goal in guided mode}
        \State $e \gets e(v,g) \in E$
        \State $u \gets \text{randomly choose from answers[e]}$
    \Else 
        \State $g \gets \text{randomly draw } g \in V \text{with }  |\text{questions[g]} | \geq 1$ \Comment{Draw new goal in free mode}
        \State $p \gets (e(v, v_1), \dots, e(v_N, g)) \subset E$
        \State $u \gets \text{randomly choose from questions[g]}$
    \EndIf

    \State $t \gets 1$
    \While{$v \neq g \wedge  t  \leq  T \wedge$ $\text{ neighbours(v)}  \neq \emptyset$  } 
    \Comment{Simulate dialog}
        \State $a \gets \text{next action from dialog policy, with inputs } v, u$ 
        \If{$a = \text{ASK(v)}$} \Comment{Print / Ask information}
            \State $\text{print } v$
            \State $e \gets e(v, v') \text{ where } v' \text{ is g in guided mode or in free mode the next node after v in the path }$
            \State $u \gets \text{randomly choose from answers[e] if type(v)} \neq  \text{information, else } \emptyset$
        \ElsIf{$a = \text{SKIP(v,v')}$} \Comment{Skip to next node}
            \State $v \gets v'$
            \State $u \gets \emptyset$
        \EndIf
        \State $r \gets \text{Calculate rewards}$
        \State $\text{Store } a, v', r \text{ in replay buffer}$
        \If{$m = \text{GUIDED}$}
            \State $g \gets \text{randomly choose neighbour(v)}$ \Comment{Draw followup goal in guided mode}
            \State $e \gets e(v,g) \in E$
        \EndIf
        \State $t \gets t + 1$
    \EndWhile
    
    \State $n \gets n + 1$
\EndFor
    
\end{algorithmic}
\end{algorithm}

\section{REIMBURSE: Dataset}
\label{appendix:dataset}

\subsection{Example User Utterances}
\label{appendix:utterances}

\subsubsection{Answer Paraphrases}
Below is an example of the answer paraphrase data we collected. The subject-area expert defined a prototypical answer to the question and then additional paraphrases were collected (through pilot study and manual expansion). Although some are very close to the prototypical answer, there can also be a lot of variation.
\newline\newline
\noindent \textbf{Prototypical Node Answer}: Mit Familie \textit{(en: With my family)}
\newline
\noindent \textbf{Associated Paraphrases}:
\begin{itemize}[noitemsep]
    \item Zusammen mit meiner Frau \textit{(en: Together with my wife)}
    \item Mit Mann und Kinder \textit{(en: With husband and kids)}
    \item Ich und meine Angeh\"origen \textit{(en: Me and my relatives)}
    \item Mit Begleitung \textit{(en: With accompaniment)}
\end{itemize}

\subsubsection{User Question Paraphrases}
For each Information node in the graph, a set of paraphrased questions were collected.
\newline\newline
\noindent \textbf{Node Text}: Eine private Verl\"angerung muss zum dienstlichen Teil der Reise verh\"altnism\"aßig sein. Das dienstliche Interesse der Reise muss im Vordergrund stehen. Bei Fragen kontaktieren Sie bitte die Reisekostenstelle. (\textit{en: A personal extension must remain in proportion with the official part of the trip. The business benefit must remain the primary focus of the trip. In case of questions, please contact the business travel department.})
\newline
\noindent \textbf{Associated Questions}:
\begin{itemize}[noitemsep]
    \item Darf ich meine Reise privat verl\"angern? \textit{(en: Can i extend my trip privately?)}
    \item Wie lang darf der private Reiseanteil sein? \textit{(en: How long is a private extension allowed to be?)}
    \item Kann ich l\"anger bleiben als der dienstliche Teil meiner Reise? \textit{(en: Can I stay longer than the official part of my trip?)}
    \item Darf ich nach meiner Reise Urlaub machen? \textit{(en: Can I add vacation to the end of my trip?)}
\end{itemize}

\subsection{Example Tree}
\label{appendix:node_types}
Below is an example of each of the node types in the dialog editor. For Information nodes (blue) subject experts could also define associated FAQ questions to help the RL agent understand user user questions related to that node.

\begin{figure}[h!]
    \centering
    \includegraphics[width=0.9\textwidth]{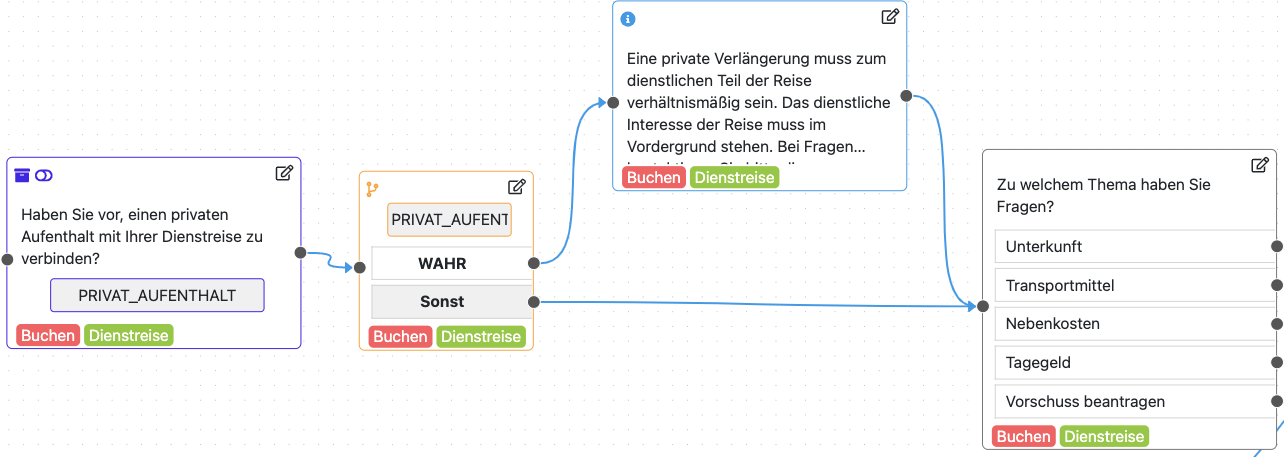}
    \caption{Example of the dialog designer interface. From left to right: Variable Nodes (purple outline), which the designer can use to ask for information relevant for later decisions, Logic Nodes (orange outline) which control the flow of the dialog based on user responses stored in the Variable Nodes (here the two options are \textit{``true''} and \textit{``else''}), Information Nodes (blue outline) where the designer can specify information which does not need a user response, and Dialog Nodes (black outline) where the designer can specify system questions and expected user responses.}
    \label{fig:dialog_designer}
\end{figure}

\begin{figure}[h!]
    \centering
    \includegraphics[width=\textwidth]{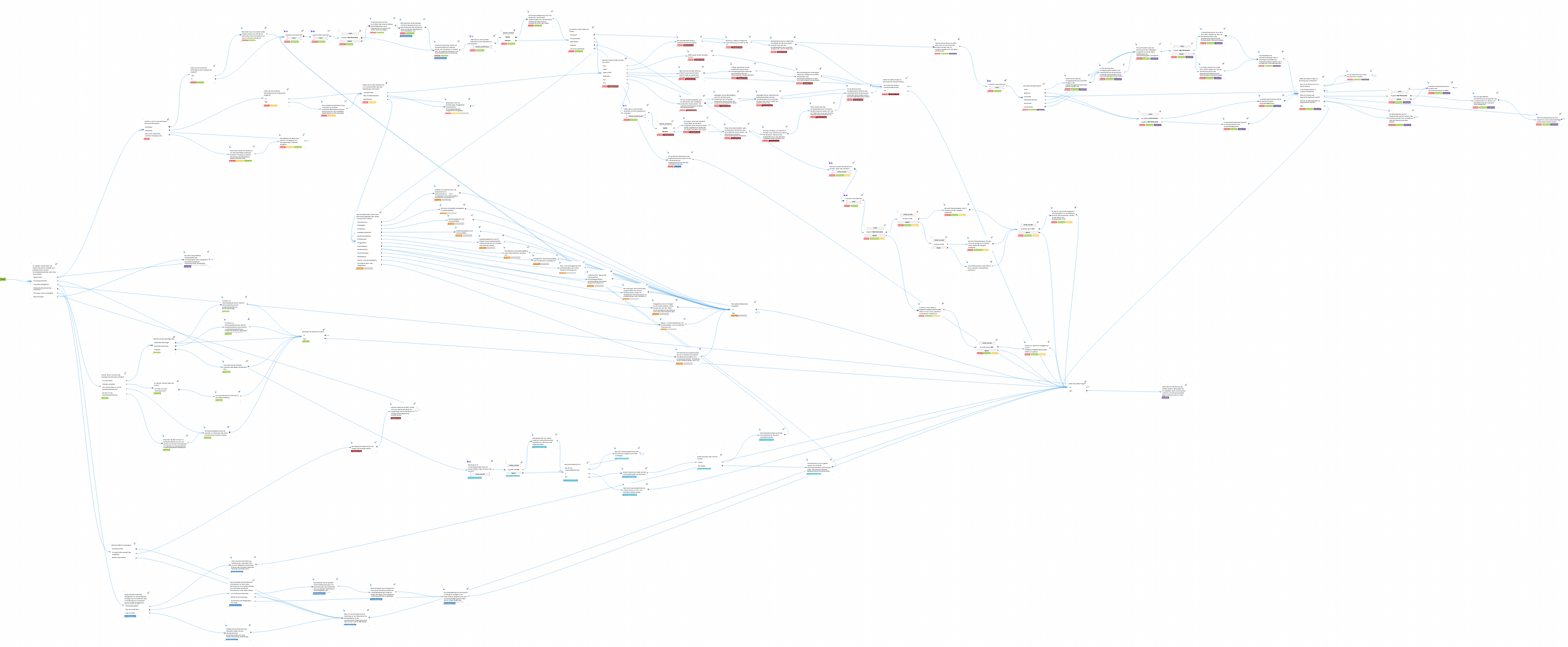}
    \caption{Overview of the dialog tree used for all experiments in this paper. The tree was designed by subject-area experts to help users better navigate the travel reimbursement process.}
    \label{fig:full_graph}
\end{figure}

\noindent Figure \ref{fig:full_graph} is a view of the full dialog tree used for experiments in this paper, consisting of 123 nodes.


\newpage
\section{Example Dialogs}
\label{appendix:dialogs}

In this section, we show parts of successful conversations between the simulator and the Sentencesim-CTS agent.

\subsection{Guided Mode}
\label{appendix:dialogs:guided}
\begin{dialogue}
    \speak{System} [ASK] Zu welchem Thema haben Sie Fragen? Sie k\"onnen entweder eine passende Option aus der Vorschlagsliste anklicken oder unten Text eingeben. (\textit{en: What topic do you have questions about? You can either click one of the suggestions or type your question.})
    \speak{User} Ein Forschungssemester planen (\textit{en:Plan a research semester})
     \speak{System} \textit{[SKIP TO] Wie lange ist ihr Forschungssemester (bitte mit Einheit angeben: Tage, Wochen oder Monate)?} (\textit{en:How long will your research semester be (please use a unit: days, weeks, or months}?)
    \speak{System} [ASK] Wie lange ist ihr Forschungssemester (bitte mit Einheit angeben: Tage, Wochen oder Monate)? (\textit{en:How long will your research semester be (please use a unit: days, weeks, or months}?)
    \speak{User} 2916 Sekunden (\textit{en: 2916 Seconds})
    \\
     \direct{
        Environment action: Match logic node branch condition $\rightarrow$ Automatic SKIP TO: Bitte wenden Sie sich zwecks möglicher Unterkünfte bei einem Aufenthalt ab 4 Wochen an die Reisekostenstelle.
    }
    \speak{System} [ASK] Bitte wenden Sie sich zwecks möglicher Unterkünfte bei einem Aufenthalt ab 4 Wochen an die Reisekostenstelle. (\textit{en: Please contact the travel department about booking your accommodations for trips longer than 4 weeks.})
    \speak{System} \textit{[SKIP TO] Liegt eine Zuweisung vor?} (\textit{en: Have you been reallocated?})
    \speak{System} [ASK] Liegt eine Zuweisung vor? (\textit{en: Have you been reallocated?})
    \\
    \direct{
        \dots CONTINUES FOR > 30 MORE TURNS
    }
\end{dialogue}

\subsection{Free Mode - Direct Skipping to Goal}
\label{appendix:dialogs:free:directskip}
\begin{dialogue}
    \speak{System} [ASK] Zu welchem Thema haben Sie Fragen? Sie k\"onnen entweder eine passende Option aus der Vorschlagsliste anklicken oder unten Text eingeben. (\textit{en: What topic do you have questions about? You can either click one of the suggestions or type your question.})
    \speak{User} Es gab ein Erdbeben, an wen soll ich mich wenden? (\textit{en: There was an earthquake, who can I contact?})
    \speak{System} \textit{[SKIP TO] Welcher Notfall ist eingetreten?} (\textit{en: What type of emergency has happened?})
    \speak{System} \textit{[SKIP TO] Was ist in einem Notfall auf einer Dienstreise zu tun? Bitte f\"uhren Sie die Notfallkarte mit der entsprechenden Notfall-Rufnummer stets mit sich: +123456789. Die Notfallkarte ist auch im Scheckkartenformat erh\"altlich. Hierzu k\"onnen Sie sich bei der Reisekostenstelle oder an ihrem Institut erkundigen. Diese Rufnummer sollte nur für Notf\"alle genutzt werden. Die Weitergabe an Dritte ist untersagt. [\dots] } (\textit{en: What should you do if an emergency occurs during a business trip? Please use the emergency card, which has the emergency telephone number +123456789. You can also get the emergency card in a platic card form. For more information, ask your institute or the business travel department. This number should only be used in emergencies. Passing it on to external people is strictly forbidden. [\dots]})
    \speak{System} [ASK] Was ist in einem Notfall auf einer Dienstreise zu tun? Bitte f\"uhren Sie die Notfallkarte mit der entsprechenden Notfall-Rufnummer stets mit sich: +123456789. Die Notfallkarte ist auch im Scheckkartenformat erh\"altlich. Hierzu k\"onnen Sie sich bei der Reisekostenstelle oder an ihrem Institut erkundigen. Diese Rufnummer sollte nur für Notf\"alle genutzt werden. Die Weitergabe an Dritte ist untersagt. [\dots] (\textit{en: What should you do if an emergency occurs during a business trip? Please use the emergency card, which has the emergency telephone number +123456789. You can also get the emergency card in a platic card form. For more information, ask your institute or the business travel department. This number should only be used in emergencies. Passing it on to external people is strictly forbidden. [\dots]})
\end{dialogue}

\subsection{Free Mode - Clarifying Question}
\label{appendix:dialogs:free:clarifying}
\begin{dialogue}
    \speak{System} [ASK] Zu welchem Thema haben Sie Fragen? Sie k\"onnen entweder eine passende Option aus der Vorschlagsliste anklicken oder unten Text eingeben. (\textit{en: What topic do you have questions about? You can either click one of the suggestions or type your question.})
    \speak{User} Muss ich etwas tun bevor ich eine Reise antreten darf? (\textit{en: Is there anything I have to do before I can travel?})
    \speak{System} \textit{[SKIP TO] Handelt es sich um eine Dienstreise oder einen Dienstgang?} (\textit{en: is this a trip within the city of employment or a trip outside of the city of employment?})
    \speak{System} [ASK] Handelt es sich um eine Dienstreise oder einen Dienstgang? (\textit{en: is this a trip within the city of employment or a trip outside of the city of employment?})
    \speak{User} Eine dienstliche Reise (\textit{en: A trip outside the city of employment})
    \speak{System} [SKIP TO] \textit{Haben Sie die schriftliche Genehmigung ihres Vorgesetzten eingeholt?} (\textit{Do you have written permission from your supervisor?})
    \speak{System} [SKIP TO] \textit{Bitte holen Sie sich zunächst mittels dieses Formulars eine schriftliche Genehmigug ihres Vorgesetzten ein. Dies ist zwingend erforderlich.} (\textit{en: Please get written permission from your supervisor as soon as possible. This is mandatory.})
    \speak{System} [ASK] Bitte holen Sie sich zunächst mittels dieses Formulars eine schriftliche Genehmigug ihres Vorgesetzten ein. Dies ist zwingend erforderlich. (\textit{en: Please get written permission from your supervisor as soon as possible. This is mandatory.})
\end{dialogue}

\newpage
\section{Training details}
\label{appendix:train_details}

\subsection{RL Model Parameters}

The following parameters were used by all user RL models (chosen through manual tuning):
\begin{table}[h!]
    \centering
    \begin{tabular}{c|c}
        \textbf{Parameter} & \textbf{Value} \\ \hline
        Layer type & Linear \\
        Activation (after each layer except in Dialog Mode Classifier Head) & SELU \\
        Shared Layer Neurons (one value / layer) & $8096, 4096, 4096$ \\
        Value Function Layer Neurons (one value / layer) & $2048, 1024$ \\
        Advantage Function Layer Neurons (one value / layer) & $4096, 2048, 1024$ \\
        Dialog Mode Classifier Neurons (one value / layer) & $256, 1$ \\
        Dropout (after each layer) & $25$\% 
    \end{tabular}
    \caption{RL Model Parameters}
\end{table}

\subsection{RL Experience Buffer}

New goals in the replay phase of the Hindsight Experience Replay are generated only for free mode dialogs.
Here, we follow the original dialog backwards until we find a suitable alternative goal node (having at least one associated user question).
We choose one of the user questions of the new goal node randomly and replace the original dialog's initial user utterance with it.
   
\begin{table}[h!]
    \centering
    \begin{tabular}{c|c}
        \textbf{Parameter} & \textbf{Value} \\ \hline
         Buffer size & $100000$ \\
         Priority Replay $\alpha$ &  $0.6$ \\
         Priority Replay $\beta$ &  $0.4$
    \end{tabular}
    \caption{RL Experience Buffer}
\end{table}

\subsection{Simulation Parameters}

The following parameters were used by our user simulator:
\begin{table}[h!]
    \centering
    \begin{tabular}{c|c}
         \textbf{Parameter} & \textbf{Value} \\ \hline
         Reward Normalization & $[-1.0, 1.0]$ \\
         Maximum Dialog Steps & $50$ \\
         User Patience & $3$ \\
         Probability guided vs. free dialog & $0.5$ \\
         Training utterance noise & $10$\% \\
    \end{tabular}
    \caption{Simulation Parameters}
\end{table}

\subsection{Dialog Mode Classifier}

We fine-tune GBERT \cite{gbert} using the Huggingface framework \cite{huggingface}.
As data, we use utterances from our train data split and evaluate on the test data split.
We have two different input pair types: 1) node text and user answer 2) node text and user FAQ question.
Training is done for 5 epochs, otherwise using the standard Huggingface \cite{huggingface} trainer class parameters.

\newpage
\subsection{RL Training Parameters}

The following parameters were used to train all RL agents (chosen through manual tuning):
\begin{table}[h!]
    \centering
    \begin{tabular}{c|c}
         \textbf{Parameter} & \textbf{Value} \\ \hline
         Optimizer & Adam \\
         Learning Rate & $1e^{-4}$ \\
         $\lambda$ & $1.0$ \\
         Maximum Training Dialog Turns & $1.5M$ \\
         Max. Gradient Norm & $1.0$ \\
         Batch Size & $128$ \\
         $\gamma$ & $0.99$ \\
         Exploration fraction of Training Turns & $0.99$ \\
         Exploration Scheme & $\epsilon$-greedy \\
         $\epsilon$ start & $0.6$ \\
         $\epsilon$ end & $0.0$ \\
         Training frequency (w.r.t. dialog turns) & $3$ \\
         Training start (w.r.t. dialog turns) & $1280$ \\
         DDQN Target Network update frequency (w.r.t. training steps) & $15$ \\
         Q-Value clipping & $10.0$ \\
         Munchausen $\tau$ & $0.03$ \\
         Munchausen $\alpha$ & $0.9$ \\
         Munchausen Clipping & $-1$ \\
         Evaluation frequency (w.r.t. dialog turns) & $10000$ \\
         Evaluation dialogs & $500$
    \end{tabular}
    \caption{RL Training Parameters}
\end{table}

\end{document}